\documentclass{bmvc2k}
\usepackage{booktabs}
\usepackage{physics}
\usepackage{mathtools}

\title{DESC: Domain Adaptation for Depth Estimation via Semantic Consistency}

\addauthor{Adrian Lopez-Rodriguez}{al4415@imperial.ac.uk}{1}
\addauthor{Krystian Mikolajczyk}{k.mikolajczyk@imperial.ac.uk}{1}

\addinstitution{
 MatchLab\\
 Imperial College London\\
 London, UK
}
\DeclarePairedDelimiter{\normsm}{\lVert}{\rVert}

\runninghead{Lopez-Rodriguez, Mikolajczyk}{DESC - Domain Adaptation for Depth Est.}

\def\eg{\emph{e.g}\bmvaOneDot}
\def\ie{\emph{i.e}\bmvaOneDot}

\def\etal{\emph{et al}\bmvaOneDot}

\begin{document}

\maketitle

\begin{abstract}
Accurate real depth annotations are difficult to acquire, needing the use of special devices such as a LiDAR sensor. Self-supervised methods try to overcome this problem by processing video or stereo sequences, which may not always be available.  Instead, in this paper, we propose a domain adaptation approach to train a monocular depth estimation model using a fully-annotated source dataset and a non-annotated target dataset. We bridge the domain gap by leveraging semantic predictions and low-level edge features to provide guidance for the target domain. We enforce consistency between the main model and a second model trained with semantic segmentation and edge maps, and introduce priors in the form of instance heights. Our approach is evaluated on standard domain adaptation benchmarks for monocular depth estimation and show consistent improvement upon the state-of-the-art.
\end{abstract}

% TODO: 
% 1-Fig. 1:  We will remove the network box, add labels Data (left) and Proposed Supervision (right), and state the semantic annotations are predicted.
% 2-Related work. Check R4
% 3-Clarify capping 50m/80m

\section{Introduction}
 State-of-the-art depth estimation methods are capable of inferring an accurate depth map from a monocular image by relying on deep learning methods that require a large amount of data with annotations~\cite{FuCVPR18-DORN, laina2016deeper}. Annotations in the form of precise depth measurements are typically provided by special tools such as a LiDAR sensor~\cite{geiger2012we} or structured light devices~\cite{silberman2012indoor}. Thus, obtaining depth annotations is costly and time-consuming. Much research has focused on developing methods not relying on directly acquired depth annotations by leveraging stereo~\cite{godard2017unsupervised, garg2016unsupervised} or video sequences~\cite{godard2019digging, casser2019struct2depth, yin2018geonet} for self-supervision. These research directions have shown promise, but a stereo pair or video sequence may not always be available in existing datasets. %% Add here motivation points.
 The use of synthetic data provides a way to obtain a large amount of accurate ground truth depth in a fast manner, however, synthetic data and real data have usually a domain gap due to the difficulty of generating photorealistic synthetic images.
 To that end, domain adaptation techniques~\cite{nath2018adadepth, zheng2018t2net} can help to transfer the models trained on an annotated source dataset $\mathcal{S}$ to a target dataset $\mathcal{T}$, reducing the burden of training a model for a new environment or camera.\par 
 Research results have shown that the domain gap for semantic segmentation and instance detection can be reduced by introducing depth information during training~\cite{liu2019synthesis, chen2019learningsemantic}. A different direction, which leverages semantic information to reduce the domain gap in depth estimation, has been less studied and mainly in multi-task scenarios~\cite{atapour2019veritatem, kundu2019adapt}. Existing datasets with semantic annotations are large and diverse in scenes as well as cameras used, hence models trained on these diverse semantic datasets are capable of generalizing to different settings~\cite{MSeg_2020_CVPR}. Several works~\cite{li2018megadepth, casser2019struct2depth} have shown that using pretrained models to obtain semantic annotations can also bring improvements in the depth estimation task. Motivated by these findings, we exploit readily-available panoptic segmentation models as guidance to bridge the gap between two different domains and to improve monocular depth estimation. 

% Semantic information can provide a guidance to depth estimation methods. Past methods support the idea that semantic information is valuable in depth estimation tasks, either posing the problem as a multi-task problem \todo{Cite papers}, using the semantic information to give extra guidance to the estimated depth \todo{Cite megadepth?}, or injecting semantic information to address the moving objects problem in video self-supervision approaches Both~\cite{li2018megadepth}, which uses semantic segmentation in their newly created dataset, and~\cite{casser2019struct2depth}, using predicted instances, use automatically extracted semantic information supporting the idea that off-the-shelf models can give valuable information without further tuning. We aim to use panoptic segmentation, including both instance detection and semantic sWe argue that semantic information can also be used to increase the performance under a domain adaptation setting. We aim to use the \textit{stuff}, i.e. the semantic maps in the panoptic setting, to form the structure of the possible depth of the image. The \textit{things}, i.e. the instances detected in the image, are used to give more precise knowledge of the depth map by using both semantic knowledge of the object and geometric information.

\begin{figure}[t]
    \centering
    \includegraphics[width=1\linewidth]{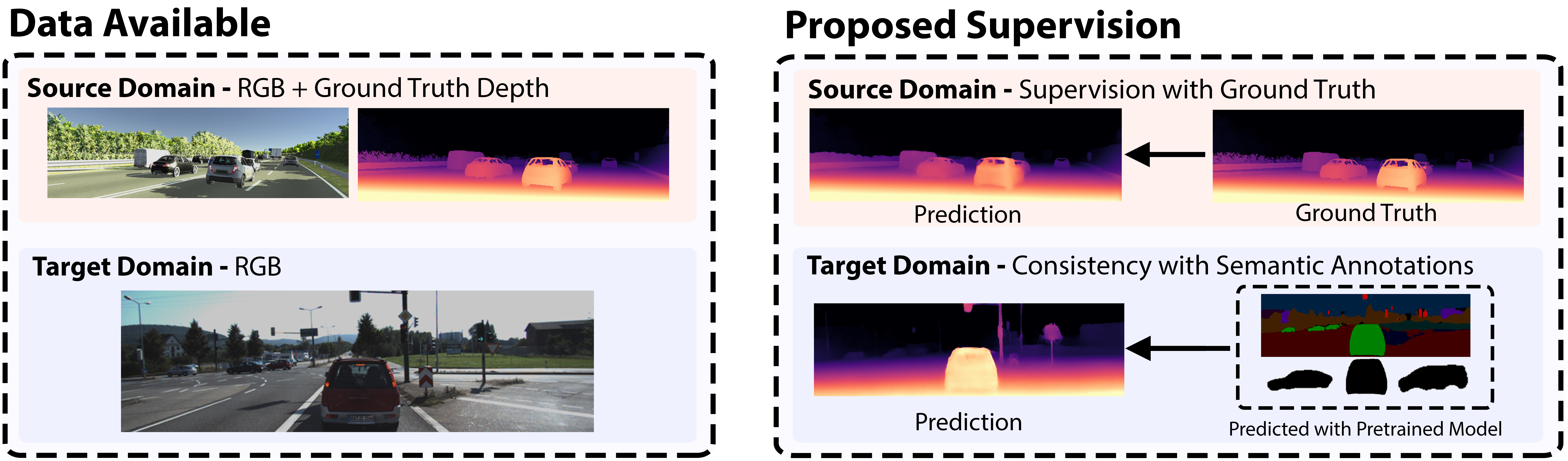}
    \caption{Overview of the data available and proposed supervision. The source domain $\mathcal{S}$ contains both RGB and ground truth depth data, and the target domain $\mathcal{T}$ contains RGB data only. We train a depth estimation model to achieve high performance in $\mathcal{T}$ by leveraging semantic annotations to introduce semantic consistency in $\mathcal{T}$. The semantic annotations are obtained using a panoptic segmentation model trained with external data.}
    \label{fig:problem}
\end{figure}

Domain adaptation approaches benefit from pseudo-labelling~\cite{chen2019progressive, saito2017asymmetric} and consistency of predictions in the source and target domains~\cite{zhao2019geometry, chen2019crdoco}. Therefore, we propose an approach that leverages semantic annotations to enforce consistency for depth estimation between the two domains, and to provide depth pseudo-labels to the target domain by using the size of the detected objects. Figure \ref{fig:problem} shows an overview of the task. Our main contributions are: (1)~the proposal of an approach to form depth pseudo-labels in the target domain by using object size priors, which are learnt in an instance-based manner in the annotated source domain; (2) the introduction of a consistency constraint with predictions from a second model trained on high-level semantics and low-level edge maps; (3) state-of-the-art results in the task of monocular depth estimation with domain adaptation from VirtualKITTI~\cite{Gaidon:Virtual:CVPR2016} to KITTI~\cite{geiger2012we}.  %Recently, using consistency-based approaches along with data-augmentation techniques have also yield an increase of performance in semi-supervised methods.\par
\section{Related Work}
\subsection{Monocular Depth Estimation}
% \noindent\textbf{Supervised Training.} Early depth estimation methods rely on supervised training, using annotations either from LiDAR~\cite{geiger2012we} or structured light scanners~\cite{silberman2012indoor}. A stack of two networks to refine the estimation from global to local was introduced in ~\cite{eigen:2014:depth}. A fully-convolutional model to regress the depth map was used in~\cite{laina2016deeper}. A strong accuracy improvement was achieved in~\cite{FuCVPR18-DORN} by discretizing the depth regression.\par
\noindent\textbf{Self-Supervision.} Early depth estimation methods rely on supervised training, using annotations from LiDAR~\cite{geiger2012we} or structured light scanners~\cite{silberman2012indoor}. Due to the difficulty of obtaining depth annotations, several works have focused on using either stereo pairs or video self-supervision. Xie \etal~\cite{xie2016deep3d} regressed a discretized disparity map and used a pixel-wise consistency loss with a second camera view, and Garg \etal~\cite{garg2016unsupervised} extended it to predict continuous depth values. The accuracy was further improved in Monodepth~\cite{godard2017unsupervised} by forcing the network to predict from a single image both left and right disparities and adding a consistency term. A stereo pair was used in Luo \etal~\cite{luo2018single} to supervise a model that synthesized the right view from the left image, and then processing both views by a stereo-matching network. Other notable approaches include the use of adversarial techniques and cycle-consistency~\cite{pilzer2019progressive, Pilzer_2019_CVPR}. Stereo images are not always available, hence video self-supervision has also been researched. Simultaneous learning of depth and pose was addressed in Zhou \etal~\cite{zhou2017unsupervised}, which given three video frames, projected the $t\text{+}1$ and $t\text{-}1$ views to the reference view $t$. Joint pose, depth and optical flow learning was proposed in GeoNet~\cite{yin2018geonet}, and Monodepth2~\cite{godard2019digging} focused on improving the pixel reprojection loss and the multi-scale loss.\par 

\noindent\textbf{Depth and Semantic Information.} Mousavian \etal~\cite{mousavian2016joint} trained a single network for both semantic and depth prediction in a multi-task manner by using a shared backbone and task-specific layers. In that direction, Chen \etal~\cite{chen2019towards} trained a network capable of selecting between depth or semantic segmentation output by only changing an intermediate task layer. In Zhang \etal~\cite{zhang2018joint} the two tasks, semantic segmentation and depth estimation, were refined alternately in a progressive manner by using a task attention module to propagate information from one task to the other. Jiao \etal~\cite{jiao2018look} proposed a novel unit to share information between the two tasks. Another method, Guizilini \etal~\cite{packnet-semguided} used a pretrained semantic segmentation network to guide the feature maps of the depth network using pixel-adaptive convolutions. In MegaDepth~\cite{li2018megadepth}, a new diverse depth dataset was collected from the internet using Multi-View Stereo and Structure-from-Motion to retrieve depth information, where semantic information was used to filter spurious depth values and to define ordinal labels. Atapour-Abarghouei and Breckon~\cite{atapour2019veritatem} assumed the availability of temporal information both in training and test time, where the different video frames were fused together to predict depth and semantic segmentation in a multi-task approach.  Struct2Depth~\cite{casser2019struct2depth}, which is more related to our work, used precomputed masks of object instances to tackle the problem of dynamic objects in video self-supervision by imposing object size constraints.
%
% \subsection{Semantic Information} 
% \todo{Not needed as there is no contribution to this area. some of it may be mentioned when discussing implementation details.}
% Semantic segmentation aims to give a class to each pixel in the image. To achieve this, fully-convolutional models were developed in~\cite{long2015fully}, and further extensions of models connecting early to latter layers directly were proposed in~\cite{ronneberger2015u, jegou2017one}. Object detection has improved dramatically its accuracy in the last few years with both two-stage detectors with a region proposal network, mainly represented by the R-CNN iterations~\cite{girshick2014rich, girshick2015fast, ren2015faster}, and single-shot detectors~\cite{redmon2016you, liu2016ssd, Redmon_2017_CVPR}.
% Panoptic segmentation~\cite{kirillov2019panoptic} is a task aiming to unify both instance detection, i.e. the \textit{things} in the image, and semantic segmentation, the \textit{stuff} in the image. For that task,~\cite{Kirillov_2019_CVPR} modified a Mask-RCNN~\cite{he2017mask} adding a segmentation branch.
%

\begin{figure}
    \centering
    \includegraphics[trim={0cm 0.8cm 0 0cm},clip,width=\linewidth]{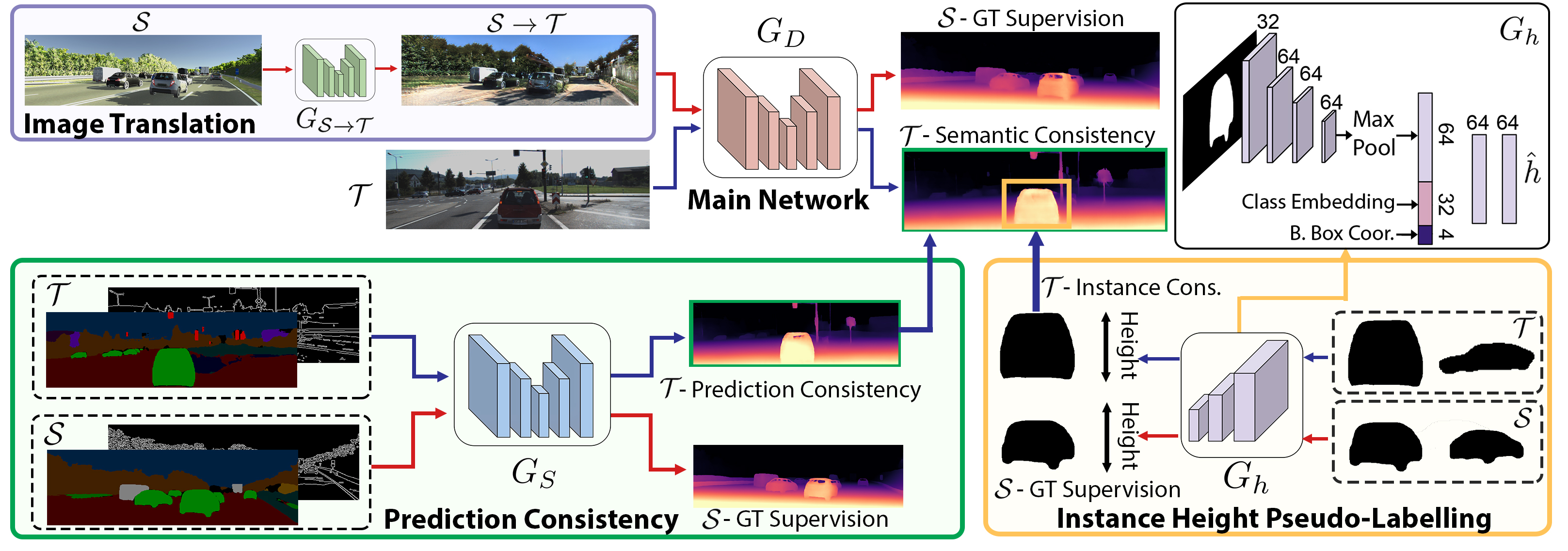}
    \caption{Overview of the approach. We train a depth estimation network $G_D$ with both target $\mathcal{T}$ and source $\mathcal{S}$ images. Source images are adapted to the style of the target images. For $\mathcal{S}$, we use ground truth supervision, while we enforce consistency with semantic information in $\mathcal{T}$. The consistency is enforced with (1) predictions from a second network $G_S$ trained with edges and semantic maps as input, and (2) depth pseudo-labels formed using an instance height $\hat{h}$ predicted by $G_h$. Both $G_S$ and $G_h$ are trained using ground truth data from $\mathcal{S}$. The architecture of $G_h$ is given in the top right. We use ReLU between the layers of $G_h$.}
    \label{fig:approach}
\end{figure}
\subsection{Domain Adaptation} 
 Domain adaptation is attracting more and more attention due to the lack of sufficient volume of annotated data for supervised training. It showed some success in areas such as classification~\cite{saito2017asymmetric, tzeng2017adversarial} and semantic segmentation~\cite{chen2019crdoco, tsai2018learning}. Popular approaches include style adaptation of the source images to match the target images~\cite{hoffman2017cycada}, enforcing consistency of predictions~\cite{roy2019unsupervised, french2018self, sajjadi2016regularization}, adversarial approaches to match either the features~\cite{ganin2015unsupervised, tzeng2017adversarial} or the outputs~\cite{ tsai2018learning} of the two domains, and using pseudo-labels~\cite{chen2019progressive, saito2017asymmetric}.\par
\noindent\textbf{Depth Estimation.} Image translation techniques have also been used for domain adaptation for depth estimation tasks~\cite{abarghouei18monocular, zhao2019geometry, zheng2018t2net, nath2018adadepth}. Atapour-Abarghouei and Breckon~\cite{abarghouei18monocular} generated synthetic data using the video-game GTA V and used a cycle-consistency approach. Additionally, during inference \cite{abarghouei18monocular} translated the target domain images to the style of the source domain before estimating the depth, adding computational burden. Our approach builds upon T$^2$Net~\cite{zheng2018t2net}, which also uses an image translation network but without a cycle-consistency loss, reducing the complexity due to the lower number of networks needed. In T$^2$Net, the target domain images are not translated during inference contrary to~\cite{abarghouei18monocular}. GASDA~\cite{zhao2019geometry} focused on the scenario where stereo supervision is available in the target domain, and added stereo photometric guidance and depth prediction consistency between original and style-transferred target domain images. GASDA~\cite{zhao2019geometry} averages during test time the depth predicted for a given target image and its corresponding style-transferred image, increasing the inference-time complexity. AdaDepth~\cite{nath2018adadepth} used an adversarial approach to align both output and feature distributions between the source and target domain, along with feature consistency to avoid mode collapse. In a multi-task setup, Kundu \etal~\cite{kundu2019adapt} developed a cross-task distillation module and contour-based content regularization to extract feature representations with greater transferability. Several synthetics datasets have been generated that can be used for depth estimation. Virtual KITTI~\cite{Gaidon:Virtual:CVPR2016} provides a synthetic version of KITTI. SYNTHIA~\cite{Ros_2016_CVPR} provides multi-camera images and depth annotations, whereas CARLA~\cite{dosovitskiy2017carla} offers a simulated environment where virtual cameras can be placed arbitrarily. In non-driving settings, some synthetic datasets that provide depth annotations are also available~\cite{MIFDB16, InteriorNet18}.\par

%%%%%%%%%%%%%%%%%%%%%
%%%%%%%%%%%%%%%%%%%%%
%%%%%%%%%%%%%%%%%%%%%
%%%%%%%%%%%%%%%%%%%%%
%%%%%%%%%%%%%%%%%%%%%
%%%%%%%%%%%%%%%%%%%%%

\section{Method}
In this section we introduce our domain adaptation for Depth Estimation via Semantic Consistency (DESC) approach. An overview is presented in Figure \ref{fig:approach}. During inference we only apply our depth estimation network $G_D$ to our target images. Semantic annotations are predicted for our source and target datasets using a panoptic segmentation model~\cite{kirillov2019panoptic} trained with external data, providing per image detected instances and a semantic segmentation map.\par

% \par
% r
% The panoptic information in $\hat{s}$ is used in our method with two different aims. First, motivated by past methods showing good performance by introducing pseudo-labels in the target domain $\mathcal{T}$, detected instances provide an approximate depth that is used pseudo-label in $\mathcal{T}$. Secondly, in this case following the literature focusing on consistency of predictions as a way to close the domain gap, the entire semantic map will be used to generate a depth map that we will use to force consistency with the main network predictions.
\subsection{Pseudo-Labelling using Instance Height}\label{sec:pl}
The height of the detected object instances can provide a strong cue for distance estimation. Struct2Depth~\cite{casser2019struct2depth} used the instance height to deal with moving objects in video self-supervision. Thus, Struct2Depth retrieved an approximate distance to the objects by solving
\begin{equation}\label{eq:distance_instance}
    \hat{D} \approx \frac{f\cdot h}{H}
\end{equation}
where $\hat{D}$ is an approximate distance to the object, $f$ is the focal length in pixels, $H$ is the predicted instance size in pixels and $h$ is the physical height of the object. It is assumed that the entire object instance is placed at a distance $\hat{D}$, that $f$ is known, and that the real object size $h$ is unknown. In Struct2Depth~\cite{casser2019struct2depth}, the object size was set as a shared learnable parameter $\hat{h}$ for the class \textit{car}, \ie, all of the detected instances of class \textit{car} were assumed to have the same height. We argue that predicting a $\hat{h}$ per object instance rather than class can provide a better height estimate, as it can take into account both intra-class variations and occlusions in the detected instances. Furthermore, instead of learning $\hat{h}$ in an unsupervised manner as in Struct2Depth~\cite{casser2019struct2depth}, we can improve the estimation using source domain data. Therefore, we use a network $G_h$, with a simple architecture presented in Figure \ref{fig:approach}, to predict a $\hat{h_i}$ for an instance $i$ from the dimensions of its bounding box, the detected binary instance mask and the predicted class label. We train $G_h$ using labels in the source data by retrieving $h_{GT,i}$, which is the ground truth physical object size for instance $i$. To retrieve $h_{GT,i}$ we use $h_{GT,i} = \frac{H_i\cdot \hat{D}_{\mathcal{S},i}}{f_\mathcal{S}}$, where the instance depth $\hat{D}_{\mathcal{S},i}$ is obtained directly from the depth ground truth. To obtain $\hat{D}_{\mathcal{S},i}$ we use $\hat{D}_i=median(M_{\mathcal{S},i}\odot y_\mathcal{S})$, where $M_{\mathcal{S},i}$ is the binary segmentation instance mask for a source domain detected instance $i$, $\odot$ refers to the Hadamard product, $y_\mathcal{S}$ is the ground truth depth, and the median operation is performed only for non-zero values. Thus, $G_h$ is trained in the source domain with $\mathcal{L}_{I,\mathcal{S}} = \frac{1}{n_I}\sum_{i}|\hat{h}_{\mathcal{S},i}-h_{GT,i}|$, where $n_I$ is the number of detected instances. In the target domain, $G_h$ is used to predict a height $\hat{h}_{\mathcal{T},i}$ for a detected instance $i$, and then $\hat{h}_{\mathcal{T},i}$ is used to retrieve a depth pseudo-label $\hat{D}_{\mathcal{T}, i}$ computed using Equation~\ref{eq:distance_instance}. We use the depth pseudo-labels $\hat{D}_{\mathcal{T}, i}$ to provide supervision for $G_D$ in the target domain using a sum of pixel-wise $L_1$ losses over all detected instances $i$,
\begin{equation}\label{eq:target_instance_loss}
\mathcal{L}_{I,\mathcal{T}} = \frac{\phi}{p_I}\sum_{i}\normsm{(\frac{\hat{D}_{\mathcal{T},i}}{\phi}-G_D(x_\mathcal{T}))\odot M_{\mathcal{T},i}}_{_1}
\end{equation}
where $p_I$ is the sum of non-zero pixels for all the binary segmentation masks $M_{\mathcal{T},i}$, $x_\mathcal{T}$ is an image from $\mathcal{T}$ and $\phi$ is a learnable scalar. The scalar $\phi$ is used to correct any scale mismatch in the predictions of $G_D(x_\mathcal{T})$ due to camera differences between $\mathcal{S}$ and $\mathcal{T}$~\cite{he2018learning}. When computing $\hat{D}_{\mathcal{T}, i}$ we use the focal length $f_\mathcal{T}$ of the target domain camera, although as we will show in Section~\ref{sec:experiments}, $\phi$ automatically scales the values to the correct range even for unknown $f_\mathcal{T}$. As we use a panoptic segmentation model trained with external data to extract semantic annotations, some of the classes detected may be present in $\mathcal{T}$ but not in $\mathcal{S}$, \eg, \textit{person} in Virtual KITTI$\rightarrow$KITTI. For those classes, $G_h$ can also learn an instance-based height prior in an unsupervised manner via consistency with $G_D$ in $\mathcal{L}_{I,\mathcal{T}}$.\par
%Due to \todo{add this} the guidance given is noisy. However, past approaches found that noisy annotations can improve the performance of a model either in a semi-supervised learning approach or in a domain adaptation setting. In this case, we aim to ..., and additionally we guide the network towards giving a similar depth to all the pixels forming an instance thus injecting some high-level constraint.\par

% An additional prior we add to the loss is that there should be a depth gradient in the borders of the instances, which holds true in most of the cases.
% \begin{equation}\label{eq:instances_edges}
%     D \approx \frac{f\cdot h}{H}
% \end{equation}
\subsection{Consistency of Predictions using Semantic Information}\label{sec:con}
Many works~\cite{roy2019unsupervised,french2018self,sajjadi2016regularization} have shown that constraining the learning process by requiring consistency in a domain adaptation setting reduces the performance gap. Similar observations have been made in semi-supervised learning~\cite{chen2020simple}, where a contrastive loss is used between different views of the same scene obtained via data augmentation. Following these findings, we enforce consistency between the predictions generated by our main depth estimation network, $G_D$, and a secondary network, $G_S$, whose input data $x_{Sem}$ is formed by two channels that have a low domain gap: a semantic segmentation map and an edge map.\par
\noindent\textbf{Semantic Structure.} A semantic segmentation map provides information on the high-level structure of the scene, and this high-level structure helps to predict the depth structure. The information is introduced in the form of an integer corresponding to the semantic class label, as we experimentally found it to yield better performance than one-hot encoding.\par 
\noindent\textbf{Edge Map.} Deep learning networks tend to use texture cues~\cite{geirhos2018imagenet} for predictions. We use an edge map to reduce the impact of the texture differences between domains, and to provide a different modality of the data to the network. Edges include information about the shapes of objects, and this shape information is valuable in depth related tasks~\cite{hu2019visualization, huang2019indoor}. Edges also present less variation and need less adaptation in domains with semantically similar scenes. \par 
\noindent\textbf{Consistency.} As both networks $G_D$ and $G_S$ receive different input modalities, forcing consistency between them for the predictions of the target domain can significantly increase the target-domain performance of both models. We propose to supervise $G_S$ with source domain depth ground truth $y_\mathcal{S}$ by using a pixel-wise $L_1$ loss, $\mathcal{L}_{Con,\mathcal{S}}$, and then force consistency of predictions in the target domain via % \begin{equation
% \mathcal{L}_{Con,\mathcal{S}} = |G_S(x_{Sem,\mathcal{S}}) - y_\mathcal{S}|
% \end{equation}
 $\mathcal{L}_{Con,\mathcal{T}}$. Then, assuming $N$ is the total number of pixels,
\begin{equation}
\mathcal{L}_{Con,\mathcal{S}} = \frac{1}{N}\norm{G_S(x_{Sem,\mathcal{S}}) - y_\mathcal{S}}_1,\;\;\;\;\;\;\    \mathcal{L}_{Con,\mathcal{T}} = \frac{1}{N} \norm{G_D(x_{\mathcal{T}}) - G_S(x_{Sem,\mathcal{T}})}_{1}
\vspace{-0.15em}
\end{equation}

\subsection{Training Loss}
We now present the modules used in DESC in addition to our semantic consistency losses.\par
\noindent\textbf{Depth Estimation Loss.} Our model $G_D$ outputs a multiscale prediction that is supervised using source domain ground truth with $\mathcal{L}_{D}$, which is a pixel-wise multiscale $L_1$ loss~\cite{zheng2018t2net, zhao2019geometry}.\par
\noindent\textbf{Image Translation.} Image translation has been demonstrated to effectively reduce the domain gap~\cite{zhao2019geometry, zheng2018t2net}. We adopt the approach from T$^2$Net~\cite{zheng2018t2net}, where a network $G_{\mathcal{S}\rightarrow \mathcal{T}}$ translates the source image to the target domain without cycle consistency. T$^2$Net~\cite{zheng2018t2net} uses a least-squares adversarial term $\mathcal{L}_{GAN}$~\cite{mao2017least} to produce examples $x_{\mathcal{S}\rightarrow\mathcal{T}}$ having a similar distribution to $x_\mathcal{T}$, and leverages the constraint imposed by $\mathcal{L}_{D}$ to ensure $x_{\mathcal{S}\rightarrow\mathcal{T}}$ is geometrically consistent with $x_{\mathcal{S}}$.  The method also uses a $L_1$ identity loss $\mathcal{L}_{IDT}=\frac{1}{N}\norm{G_{\mathcal{S}\rightarrow \mathcal{T}}(x_\mathcal{T})-x_\mathcal{T}}_1$ to force $G_{\mathcal{S}\rightarrow \mathcal{T}}(x_\mathcal{T})\approx x_\mathcal{T}$, \ie, $\mathcal{L}_{IDT}$ forces $G_{\mathcal{S}\rightarrow \mathcal{T}}$ to behave as an identity mapping for $x_\mathcal{T}$.
% A reconstruction loss is also added to force $G_{S\rightarrow T}$ to function approximately as an identity function for any given target image by using
% \begin{equation}
%     \mathcal{L}_{Rec} = |G_{S\rightarrow T}(x_{\mathcal{T}})-x_{\mathcal{T}}|
% \end{equation} 
\par
% \begin{equation}
%     \mathcal{L}_{\mathcal{S}} = |G_D(x_{\mathcal{S}})-y_{\mathcal{S}}|
% \end{equation}
\noindent\textbf{Smoothing.} We use for the target data the smoothing term $\mathcal{L}_{Sm}$ introduced in Monodepth~\cite{godard2017unsupervised}, and successfully used in domain adaptation~\cite{zheng2018t2net,zhao2019geometry} methods for depth estimation.\par
% \begin{equation}
%     \mathcal{L}_{Sm} = |\partial_x G_D(x_R)|e^{-|\partial_x x_R|} +  |\partial_y G_D(x_R)|e^{-|\partial_y x_R|}
% \end{equation}
\noindent\textbf{Overall Loss.} Our final model is trained using the following loss
\begin{equation}
    \mathcal{L} = \lambda_\mathcal{S} (\mathcal{L}_\mathcal{D} + \mathcal{L}_{Con,\mathcal{S}} + \mathcal{L}_{I,\mathcal{S}}) +  \lambda_\mathcal{T} (\mathcal{L}_{Con,\mathcal{T}} + \mathcal{L}_{I,\mathcal{T}}) + \lambda_{Sm} \mathcal{L}_{Sm} + \lambda_{IDT} \mathcal{L}_{IDT} + \lambda_{GAN} \mathcal{L}_{GAN}
\end{equation}
where $\lambda_\mathcal{S}, \lambda_\mathcal{T}, \lambda_{Sm}, \lambda_{IDT}, \lambda_{GAN}$ are hyperparameters to balance the different terms. 
\begin{table}\scriptsize\centering
\begin{tabular}{ lccccccc }\toprule
\multicolumn{1}{c}{} & \multicolumn{4}{c}{Lower is better} &   \multicolumn{3}{c}{Higher is better}\\
\cmidrule(lr){2-5}\cmidrule(lr){6-8}
Method & Abs Rel & Sq Rel & RMSE & RMSE log & $\delta<1.25$ & $\delta<1.25^2$ & $\delta<1.25^3$\\ \midrule
\textbf{Cap 80m} &   &  &  &  &  &  &  \\
AdaDepth~\cite{nath2018adadepth} & 0.214 & 1.932 & 7.157 & 0.295 & 0.665 & 0.882 & 0.950 \\
T$^2$Net~\cite{zheng2018t2net} & 0.173  &   1.396  &   6.041  &   0.251  &   0.757  &   0.916  &   0.966\\
DESC & \textbf{0.156}  &   \textbf{1.067}  &   \textbf{5.628}  &   \textbf{0.237}  &   \textbf{0.787 } &   \textbf{0.924}  &   \textbf{0.970}  \\
% Img-St. Tr.~\cite{abarghouei18monocular} & Syn & K+G & 0.110 & 0.929 & 4.726 & 0.194 & 0.923 & 0.967 & 0.984\\
\midrule
AdaDepthS~\cite{nath2018adadepth} &  0.167 & 1.257 & 5.578 & 0.237 & 0.771 & 0.922 & 0.971\\
\midrule
\textbf{Cap 50m} &  &  &  &  &  &  & \\
AdaDepth~\cite{nath2018adadepth}  &  0.203 & 1.734 & 6.251 & 0.284 & 0.687 & 0.899 & 0.958 \\
T$^2$Net~\cite{zheng2018t2net} &  0.165  &   1.034  &   4.501  &   0.235  &   0.772  &   0.927  &   0.972 \\
DESC &  \textbf{0.149}  &   \textbf{0.819}  &   \textbf{4.172}  &   \textbf{0.221}  &   \textbf{0.805}  &   \textbf{0.934}  &   \textbf{0.975} \\
\midrule
AdaDepthS~\cite{nath2018adadepth}  & 0.162 & 1.041 & 4.344 & 0.225 & 0.784 & 0.930 & 0.974 \\
\bottomrule
\end{tabular}
\caption{Results for Virtual KITTI$\rightarrow$KITTI in KITTI~\cite{geiger2012we} Eigen~\cite{eigen:2014:depth} split. Results from \textit{T$^2$Net} are recomputed using median scaling and the official pretrained model. \textit{AdaDepthS} is a semi-supervised method using additionally 1000 annotated KITTI images for training.}
\label{tab:results_kitti}
\end{table}

\section{Experiments}\label{sec:experiments}
We discuss the experimental setup before presenting our evaluation results.\par 
\noindent\textbf{Setup.} We use Pytorch 1.4 and an NVIDIA 1080TI GPU. We obtain the semantic annotations in both $\mathcal{S}$ and $\mathcal{T}$, by using a ResNet-101~\cite{he2016deep} panoptic segmentation model~\cite{Kirillov_2019_CVPR} trained in COCO-Stuff~\cite{lin2014microsoft, caesar2018coco} from the \textit{Detectron 2} library~\cite{wu2019detectron2}. We employ a U-Net~\cite{ronneberger2015u} for $G_D$ and $G_{S}$, and a ResNet-based model for $G_{\mathcal{S}\rightarrow\mathcal{T}}$. Both image translation and depth estimation architectures are the same as the architectures used in~\cite{zheng2018t2net, zhao2019geometry}. Following~\cite{zhao2019geometry}, we set $\lambda_\mathcal{S}=50$,  $\lambda_{GAN}=1$, $\lambda_{Sm}=0.01$, and following~\cite{zheng2018t2net} we set $\lambda_{IDT}=100$. Similarly to the original implementation of~\cite{zhao2019geometry}, we first pretrain the networks to reach good performance in $\mathcal{S}$ before introducing the consistency terms, \ie, with $\lambda_\mathcal{T}=0$. Afterwards, we freeze $G_{\mathcal{S}\rightarrow\mathcal{T}}$ to reduce the memory footprint, and we introduce the semantic consistency terms by setting $\lambda_\mathcal{T}=1$ unless stated otherwise. The batch size is set to 4, with a 50/50 target and source data ratio, we use Adam~\cite{kingma2015adam} with learning rate $10^{-4}$ and we train for 20,000 iterations after pretraining. To obtain the edge map for $G_{S}$ we use a Canny Edge detector~\cite{canny1986computational}. We randomly change the brightness, saturation and contrast of the images for data augmentation.\par
\begin{table}[t]\scriptsize\centering
\begin{tabular}{ lccccccc }\toprule
\multicolumn{1}{c}{} & \multicolumn{4}{c}{Lower is better} &   \multicolumn{3}{c}{Higher is better}\\
\cmidrule(lr){2-5}\cmidrule(lr){6-8}
Method & Abs Rel & Sq Rel & RMSE & RMSE log & $\delta<1.25$ & $\delta<1.25^2$ & $\delta<1.25^3$ 
\\\midrule
Only Source~\cite{zhao2019geometry} & 0.223  &   2.205  &   7.055  &   0.305  &   0.672  &   0.872  &   0.945\\
+ \text{Img.} & 0.199  &   2.436  &   7.137  &   0.280  &   0.753  &   0.890  &   0.950 \\ 
+ \text{Img.} + Con. (only edges) & 0.187  &   1.330  &   6.094  &   0.258  &   0.708  &   0.905  &   0.966\\ 
+ \text{Img.} + Con. & 0.173  &  1.235  &   5.776  &   0.244  &   0.748  &   0.919  &   0.969 \\ 
+ \text{Img.} + Ins. & 0.171  &   1.332  &   5.818  &   0.250  &   0.771  &   0.918  &   0.966 \\ 
+ \text{Img.} + Ins. ($\lambda_{Sm}=0.1$) & 0.165  &   1.157  &   5.670  &   0.245  &   0.774  &   0.921  &   0.968 \\ 
DESC - Full (1 $h$ per class~\cite{casser2019struct2depth})  & 0.160  &   1.107  &   5.746  &   0.243  &   0.780  &   0.920  &   0.968\\
DESC - Full (unknown $f_\mathcal{T}$) & \textbf{0.156}  &   1.084  &   5.654  &   \textbf{0.237}  &   0.783  &   \textbf{0.926}  & \textbf{0.971} \\
DESC - Full & \textbf{0.156}  &  \textbf{1.067}  &   \textbf{5.628}  &   \textbf{0.237}  &   \textbf{0.787}  &   0.924  &   0.970\\
\midrule
$G_S$ & 0.186  &   2.164  &   7.011  &   0.282  &   0.763  &   0.894  &   0.949 \\ 
\bottomrule
 \hline
\end{tabular}
\caption{Ablation study of DESC for Virtual KITTI$\rightarrow$KITTI in Eigen split~\cite{eigen:2014:depth} capped at 80m. \textit{Img.} refers to using image translation, \textit{Ins.} to using instance-height pseudo-labels (Section \ref{sec:pl}) and \textit{Con.} to the consistency of predictions constraint (Section \ref{sec:con}).}
\label{tab:ablation}
\end{table}

\noindent\textbf{Virtual KITTI$\rightarrow$KITTI.} We follow the same experimental settings as in~\cite{zheng2018t2net, zhao2019geometry}. Both Virtual KITTI~\cite{Gaidon:Virtual:CVPR2016} and KITTI~\cite{geiger2012we} images are downscaled to 640x192, and following~\cite{zheng2018t2net} we cap the Virtual KITTI~\cite{Gaidon:Virtual:CVPR2016} ground truth depth at 80m.\par
\noindent\textbf{Cityscapes$\rightarrow$KITTI.} Cityscapes~\cite{cordts2016cityscapes} provides disparity maps computed using Semi-Global Matching~\cite{hirschmuller2007stereo}. We use the official training set, consisting of 2975 images of size 2048x1024. We set the horizon line approximately in the center by cropping the upper part, resulting in images of 2048x964. We then take the 2048x614 center crop to have the same aspect ratio as in KITTI and rescale the images to 640x192. We use $\lambda_\mathcal{T}=5$ for this experiment.\par
%\noindent\textbf{SYNTHIA to Kitti} The frontal images from the right camera in Synthia for the spring season were used for this experiment. KITTI follows the same settings as in the VirtualKITTI to KITTI setup. Here compare to GASDA without geometric.\par
% \noindent\textbf{InteriorNet to NYU} In~\cite{adadepth} the adaptation was performed from SUNCG? to NYUv2, however due to licensing issues SUNCG is not currently available online. For our experiments in indoors data we perform adaptation from a subset of InteriorNet to NYU v2. The size of the images is 640x192. Here compare to GASDA without geometric.
\noindent\textbf{Evaluation in KITTI}. We follow the same evaluation protocol, metrics and splits as in Eigen \etal~\cite{eigen:2014:depth} for KITTI, using the evaluation code from Monodepth2~\cite{godard2019digging}. The predictions are upscaled to match the ground truth size. The results are reported using median scaling as in past methods~\cite{nath2018adadepth, casser2019struct2depth, zhou2017unsupervised}, except when using stereo supervision in KITTI. We provide results for both ground truth depth capped at 80m and between 1-50m as done in~\cite{zhao2019geometry,zheng2018t2net}.

\subsection{Quantitative Results}
\noindent\textbf{Comparison with State-of-the-Art.} Table \ref{tab:results_kitti} compares the performance of DESC with the Virtual KITTI$\rightarrow$KITTI state-of-the-art methods not using stereo nor video self-supervision in KITTI. DESC performs better than AdaDepth~\cite{nath2018adadepth} and T$^2$Net~\cite{zheng2018t2net}, with a Sq. Rel. error almost $24\%$ lower than T$^2$Net.  We also include \textit{AdaDepthS}, which is a version of AdaDepth that uses 1000 annotated KITTI images for training in addition to Virtual KITTI ground truth. We improve upon \textit{AdaDepthS} in most metrics without using any KITTI annotations.\par
%The domain adaptation method from~\cite{abarghouei18monocular} is also in Table \ref{tab:results_kitti}, although the source domain dataset was formed by images from GTA5 and is not publicly available. Compared to Virtual Kitti, as argued in~\cite{zhao2019geometry} their GTA5 dataset has a lower domain gap due to more photo-realistic images and is more than three times larger as it contains 70000 training images. Due to this disparity between the datasets, the results from~\cite{abarghouei18monocular} are not directly comparable to the results using Virtual Kitti. 
%The reason for the better performance in squared metrics is the structure of the scene given by the consistency with $G_s$.\par

\begin{table}\scriptsize\centering
\begin{tabular}{ lcccccccc}\toprule
\multicolumn{1}{c}{} & \multicolumn{4}{c}{Lower is better} &   \multicolumn{3}{c}{Higher is better}\\
\cmidrule(lr){2-5}\cmidrule(lr){6-8}
Method & Abs Rel & Sq Rel & RMSE & RMSE log & $\delta<1.25$ & $\delta<1.25^2$ & $\delta<1.25^3$\\ \midrule
\textbf{Virtual KITTI$\rightarrow$ KITTI}\\
Source + Stereo & 0.131  &   1.154  &   5.518  &   0.227  &   0.837  &   0.937  &   0.971 \\
T$^2$Net~\cite{zheng2018t2net} + Stereo & 0.126  &   1.114  &   5.429  &   0.223  &   0.839  &   0.938  &   0.971\\
GASDA~\cite{zhao2019geometry} & 0.124  &   1.018  &   5.202  &   \textbf{0.217}  &   \textbf{0.846}  &   \textbf{0.944}  &   0.973\\
DESC + Stereo & \textbf{0.119}  &   \textbf{0.935}  &  \textbf{5.050}  &   \textbf{0.217}  &   0.843  &   0.942  &   \textbf{0.974}  \\
\midrule
\textbf{Only KITTI}\\
Monodepth2 (w/o pre.)~\cite{godard2019digging} & 0.130 & 1.144 & 5.485 & 0.232 & 0.831 & 0.932 & 0.968 \\ 
Monodepth2 (ImageNet pre.)~\cite{godard2019digging}  & 0.109 & 0.873 & 4.960 & 0.209 & 0.864 & 0.948 & 0.975\\ 

\bottomrule
\end{tabular}
\caption{Results in KITTI Eigen split (80m cap) for methods using stereo data in KITTI. Due to an evaluation error in \cite{zhao2019geometry}, results from GASDA are recomputed using the official pretrained models. We include the state-of-the-art stereo-trained method \textit{Monodepth2}~\cite{godard2019digging}.}
\label{tab:results_kitti_stereo}
\end{table}

\noindent\textbf{Ablation Study.} Table \ref{tab:ablation} shows an ablation study of DESC. The result marked with $+Img$ correspond to T$^2$Net~\cite{zheng2018t2net} without the adversarial feature module, and with a lower smoothing weight $\lambda_{Sm}$ as we use  $\lambda_{Sm}=0.01$ instead of the $\lambda_{Sm}=0.1$ used for the T$^2$Net implementation shown in Table \ref{tab:results_kitti}. The lower $\lambda_{Sm}$ we use accounts for the better results of T$^2$Net in Table \ref{tab:results_kitti}. We chose a smaller $\lambda_{Sm}$ for our experiments because a larger $\lambda_{Sm}$ blurs the predictions, leading to a worse result after enforcing consistency with $G_S$ due to the loss of detail. However, a larger $\lambda_{Sm}$ is beneficial when consistency with $G_S$ is not applied as shown by the improved results of \textit{+Img.+Ins. ($\lambda_{Sm}=0.1$)} compared to \textit{+Img.+Ins.}. Both the instance-based pseudo-labelling and consistency with $G_S$ modules bring an improvement as shown in \textit{+Img.+Ins.} and \textit{+Img.+Con.} compared to \textit{+Img}. Using the consistency term in the case where only edge maps are input into $G_S$ improves most metrics as shown in \textit{+ Img.+Con. (only edges)}, although it also shows that inputting the semantic map into $G_S$ is largely beneficial. We argue that the better results of \textit{+Img.+Con.} compared to \textit{Img} are not due to a distillation process, \ie, due to $G_S$ having a higher accuracy than $G_D$ after source data pretraining. Table~\ref{tab:ablation} shows in the line $G_S$ the accuracy when evaluating $G_S$ after source data pretraining (\ie, before $G_D$ consistency), and its lower performance compared to \textit{+Img.+Con.} suggests that consistency is the reason for the accuracy increase. \textit{DESC - Full} shows an improvement in all metrics, also compared to learning a single $h$ per class as in Struct2Depth~\cite{casser2019struct2depth}. For \textit{DESC - Full (unknown $f_\mathcal{T}$)} we set $f_\mathcal{T}$ to half the actual value, obtaining comparable results to when using the correct value of $f_\mathcal{T}$, \ie, in \textit{DESC - Full}. This result shows that $\phi$ in Equation~\ref{eq:target_instance_loss} automatically scales the instance size pseudo-labels to the correct range for unknown $f_\mathcal{T}$.\par
% \noindent\textbf{Quality of weak annotations.} We now analyze the accuracy of the weak labels obtained via. To do so, we compare the weak depth in the annotation with the ground-truth in those areas in Table \ref{fig:}. We also include the accuracy of the model trained only with image transfer for those regions. We show ... In Figure \ref{fig:pl} we analyze the frequency of the found instances in the training set of both KITTI and Virtual KITTI, along with the estimated height for the instances, and the mean and std of the relative error in KITTI per class. Compared to the constrain for video supervision in~\cite{casser2019struct2depth}, we obtain the height prior from the source domain instead of using a learnable parameter. We show in Table \ref{tab:ablation} the obtained performance when we follow the same method.\par

 \noindent\textbf{Stereo Supervision.} Although DESC focuses on the setting where no self-supervision is used in $\mathcal{T}$, our approach can also bring an improvement in such a scenario.  We train DESC adding stereo supervision in KITTI by adding the same multiple-scale pixel-wise reconstruction method as in GASDA~\cite{zhao2019geometry} with the same loss weight of $\lambda_{St}=50$. To account for the introduced supervision in $\mathcal{T}$, we increase $\lambda_\mathcal{T}=5$ and the number of training iterations to 100,000. Table \ref{tab:results_kitti_stereo} shows that, compared to \textit{T$^2$Net+Stereo}, our method with stereo supervision, \textit{DESC + Stereo}, achieves better results in all metrics and also outperforms GASDA~\cite{zhao2019geometry} in most metrics. GASDA is a domain adaptation method tailored for stereo supervision that uses two depth estimation networks and an image-translation network during inference. We report better performance than the state-of-the-art for stereo supervision, Monodepth2~\cite{godard2019digging} without ImageNet~\cite{imagenet_cvpr09} pretraining in \textit{Monodepth2 (w/o pre.)}. However, ImageNet pretraining has a positive effect on the accuracy, shown in  \textit{Monodepth2 (ImageNet pre.)}.

% \begin{table}\scriptsize\centering
% \begin{tabular}{ llllllllll }\toprule
% Method & Sup. & Abs Rel & Sq Rel & RMSE & RMSE log & $\delta<1.25$ & $\delta<1.25^2$ & $\delta<1.25^3$ 
% \\\midrule
% $G_S$ (E) & Syn &  \\ 
% $G_S$ (S+E) & Syn & 0.164  &   1.310  &   5.853  &   0.257  &   0.772  &   0.906  &   0.960 \\ 
% $G_S$ (S**+E) & Syn &  &    \\ 
% $G_S$ (I+E) & Syn &  \\
% $G_S$ (S+I) & Syn &  \\
% $G_S$ (S+I+E) & Syn & 0.164  &   1.281  &   5.797  &   0.251  &   0.774  &   0.909  &   0.963 \\ 
% \bottomrule
%  \hline
% \end{tabular}
%  \textbf{Legend:} \emph{K}-KITTI, \emph{VK}-Virtual KITTI, \emph{G}-GTA,  \emph{St}-Stereo, \emph{Syn}-Synthetic
% \caption{Ablation study of the proposed method}
% \label{tab:results_kitti}
% \end{table}

\begin{table}\scriptsize\centering
\begin{tabular}{ lcccccccc }\toprule
\multicolumn{1}{c}{} & \multicolumn{4}{c}{Lower is better} &   \multicolumn{3}{c}{Higher is better}\\
\cmidrule(lr){2-5}\cmidrule(lr){6-8}
Method &  Abs Rel & Sq Rel & RMSE & RMSE log & $\delta<1.25$ & $\delta<1.25^2$ & $\delta<1.25^3$ 
\\\midrule
\textbf{Virtual KITTI$\rightarrow$KITTI}\\
T$^2$Net~\cite{zheng2018t2net} & 0.151  &   1.535  &   6.177  &   0.224  &   0.817  &   0.935  &   0.975 \\
DESC & 0.120  &   0.968  &   5.597  &   0.206  &   0.839  &   0.937  &   0.977   \\
GASDA~\cite{zhao2019geometry} & 0.095  &   1.068  &   5.015  &   0.168  &   0.906  &   0.966  &   \textbf{0.986} \\
DESC + Stereo & \textbf{0.085}  &   \textbf{0.781}  &  \textbf{4.490}  &   \textbf{0.158}  &   \textbf{0.909}  &   \textbf{0.967}  &   \textbf{0.986}  \\
\midrule
\textbf{Only KITTI}\\
Monodepth2 (w/o pre.)~\cite{godard2019digging} & 0.096  &   1.163  &   5.161  &   0.179  &   0.898  &   0.959  &   0.981 \\
Monodepth2 (ImageNet pre.)~\cite{godard2019digging} & 0.082  &   0.908  &   4.698  &   0.158  &   0.919  &   0.970  &   0.986 \\
\bottomrule
 \hline
\end{tabular}
\caption{Results on the KITTI 2015 stereo 200 training set disparity images~\cite{menze2015object, geiger2012we}. We include \textit{Monodepth2}~\cite{godard2019digging}, the state-of-the-art stereo method trained only in KITTI. }
\label{tab:results_stereo_data}
\end{table}

% \begin{table}\scriptsize\centering
% \begin{tabular}{ lllll }\toprule
% Method & Sup. & Abs Rel & Sq Rel & RMSE
% \\\midrule
% Karsch et al. & GT & 0.398 & 4.723 & 7.801\\
% Laina et al.  & GT & 0.198 & 1.665 & 5.461\\
% Kundu et al.  & GT & 0.452 & 5.71 & 9.559\\
% \midrule
% Godard et al.  & No & 0.505 & 10.172 & 10.936\\
% Kundu et al.  & No & 0.647 & 12.341 & 11.567\\
% Atapour et al.  & No & 0.423 & 9.343 & 9.002\\
% GASDA & No & 0.403 & 6.709 & 10.424\\
% Ours & No &  - & - & -\\
% Ours - $G_S$ & No &  - & - & -\\ 
% Ours + St. & No &  - & - & -\\

% \bottomrule
%  \hline
% \end{tabular}
% \caption{Evaluation on Make3D.}
% \label{tab:results_stereo}
% \end{table}
\begin{table}\scriptsize\centering
\begin{tabular}{ lcccccccc }\toprule
\multicolumn{1}{c}{} & \multicolumn{4}{c}{Lower is better} &   \multicolumn{3}{c}{Higher is better}\\
\cmidrule(lr){2-5}\cmidrule(lr){6-8}
Method & Abs Rel & Sq Rel & RMSE & RMSE log & $\delta<1.25$ & $\delta<1.25^2$ & $\delta<1.25^3$ 
\\\midrule
\textbf{Only Cityscapes}\\
Source Baseline & 0.189  &  1.717  &   6.478  &   0.257  &   0.740  &   0.919  &   0.968\\
Struct2Depth (M)~\cite{casser2019struct2depth} & 0.188 & 1.354 & 6.317 & 0.264 & 0.714 & 0.905 & 0.967\\
Struct2Depth (M+R)~\cite{casser2019struct2depth} & 0.153 & 1.109 & 5.557 & 0.227 & 0.796 & 0.934 & 0.975\\
\midrule
\textbf{Cityscapes$\rightarrow$ KITTI}\\
T$^2$Net~\cite{zheng2018t2net} & 0.173  &   1.335  &   5.640  &   0.242  &   0.773  &   0.930  &   0.970 \\
DESC (Img.+Ins.) & 0.174  &   1.480  &   5.920  &   0.240  &   0.782  &   0.931  &   0.971\\
DESC (Img.+Con.) &  0.150  &   0.981  &   5.359  &   \textbf{0.222}  &   0.805  &   0.938  &   \textbf{0.976}\\
DESC (Full, $\phi=1$) & 0.169  &   1.142  &   5.936  &   0.261  &   0.741  &   0.919  &   0.967 \\
DESC (Full, $\phi$ learnt) & \textbf{0.149}  &   \textbf{0.967}  &   \textbf{5.236}  &   0.223  &   \textbf{0.810}  &   \textbf{0.940}  &   \textbf{0.976}\\

\bottomrule
 \hline
\end{tabular}
\caption{Cityscapes$\rightarrow$KITTI results, evaluated in KITTI~\cite{geiger2012we} Eigen split (80m cap). \textit{Struct2Depth (M+R)}~\cite{casser2019struct2depth} uses three consecutive frames for refinement.}
\label{tab:cityscapes_kitti}
\end{table}
\noindent\textbf{Evaluation on KITTI Stereo.} KITTI Stereo 2015~\cite{menze2015object} provides images annotated in a process combining (1) static background retrieval via egomotion compensation and (2) fitting of CAD models to account for dynamic objects. The result is a denser ground truth compared to the LiDAR depth annotations provided in KITTI, especially in the cars. DESC, which uses instances pseudo-labels, benefits from evaluating in images with denser annotation in the vehicles, as shown in Table \ref{tab:results_stereo_data} in the larger accuracy gap between \textit{DESC} and \textit{T$^2$Net}, and also between \textit{DESC + Stereo} and \textit{GASDA} compared to Table \ref{tab:results_kitti} and Table \ref{tab:results_kitti_stereo}. \textit{DESC + Stereo} achieves either better (Sq Rel, RMSE) or equal (RMSE log) squared metrics results than the state-of-the-art \textit{Monodepth2 (ImageNet pre.)} without pretraining $G_D$ in ImageNet.\par
% \noindent\textbf{Make3D Evaluation.} Similarly to~\cite{zhao2019geometry}, we evaluate our method in Make3D~\cite{saxena2006learning, saxena2008make3d}. However, contrary to~\cite{zhao2019geometry}, we do not require stereo information in the target domain, hence we can finetune in Make3D using our method for the 400 training images available.\par 
\noindent\textbf{Cityscapes$\rightarrow$ KITTI.} Table \ref{tab:cityscapes_kitti} shows the results for this benchmark. We improve upon T$^2$Net for all metrics, with a 13.9\% lower absolute relative error. Most of the accuracy improvement comes from the consistency term as shown in \textit{DESC (Img.+Con.)} and \textit{DESC (Full, $\phi$ learnt)}. Due to the camera difference between the datasets, the learnable scalar $\phi$ is necessary for good performance, as shown for fixed $\phi=1$ in \textit{DESC (Full, $\phi$=1)}. Struct2Depth~\cite{casser2019struct2depth} also uses precomputed semantic annotations to improve its self-supervised video learning, although Struct2Depth is not a domain adaptation method as it only trains with Cityscapes~\cite{cordts2016cityscapes} data, \ie, it does not use KITTI for training. Struct2Depth also uses a different crop for Cityscapes. Table \ref{tab:cityscapes_kitti} shows that we achieve better accuracy than \textit{Struct2Depth (M+R)}, which uses three frames at test time for refinement, whereas we only need a single image for inference.
\begin{figure}
    \centering
    \includegraphics[trim={0cm 6.4cm 0 6cm},clip, width=\linewidth]{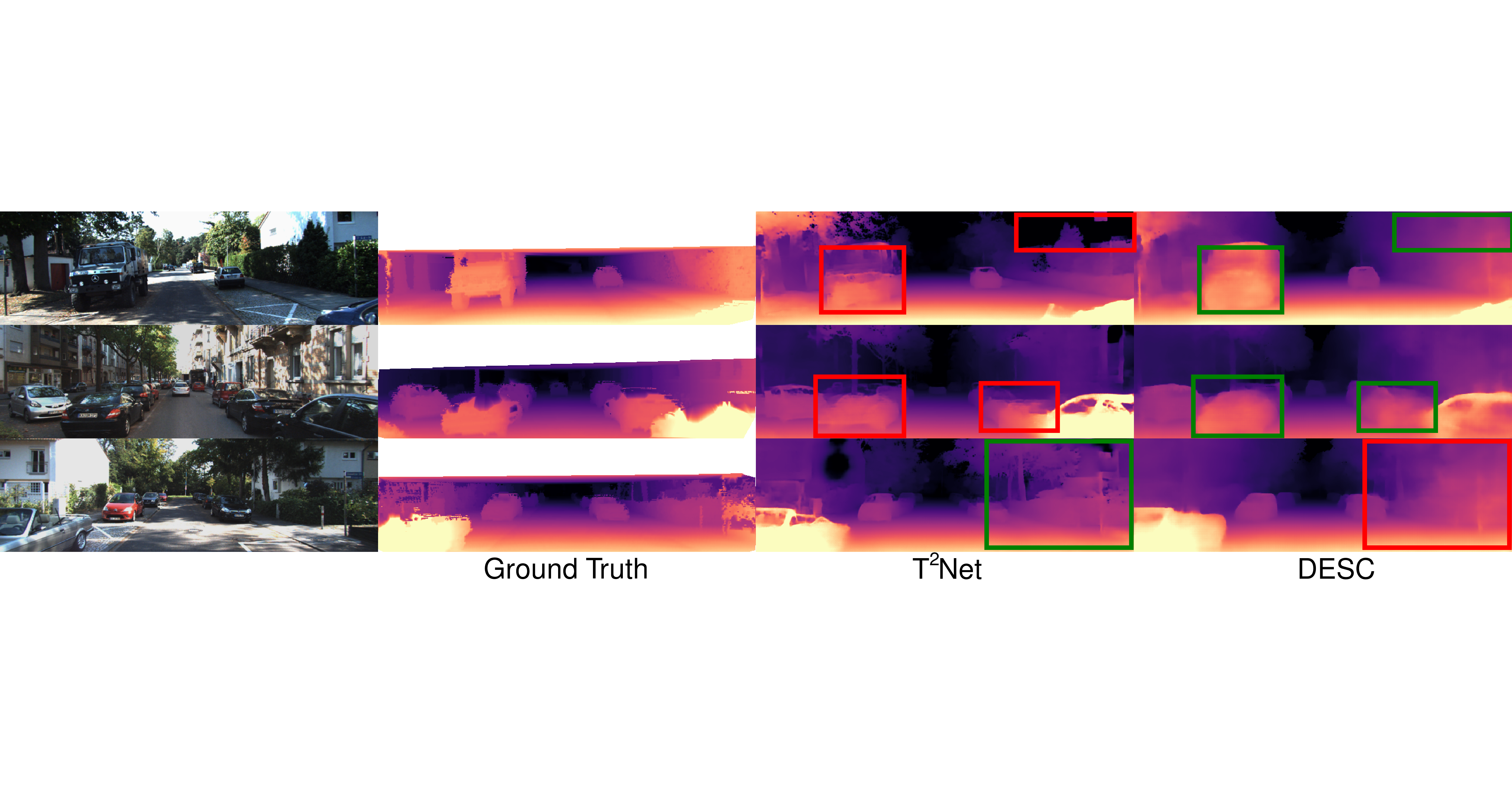}
    \vspace{-1.8em}
    \caption{Qualitative results in KITTI for models trained in Virtual KITTI$\rightarrow$KITTI. Ground truth depth is linearly interpolated for visualization. Green bounding boxes refer to areas of the prediction more accurate compared to the corresponding red bounding boxes.}
    \label{fig:qual_res}
\end{figure}
\begin{figure}
    \centering
    \includegraphics[trim={0cm 5.72cm 0 6cm},clip, width=\linewidth]{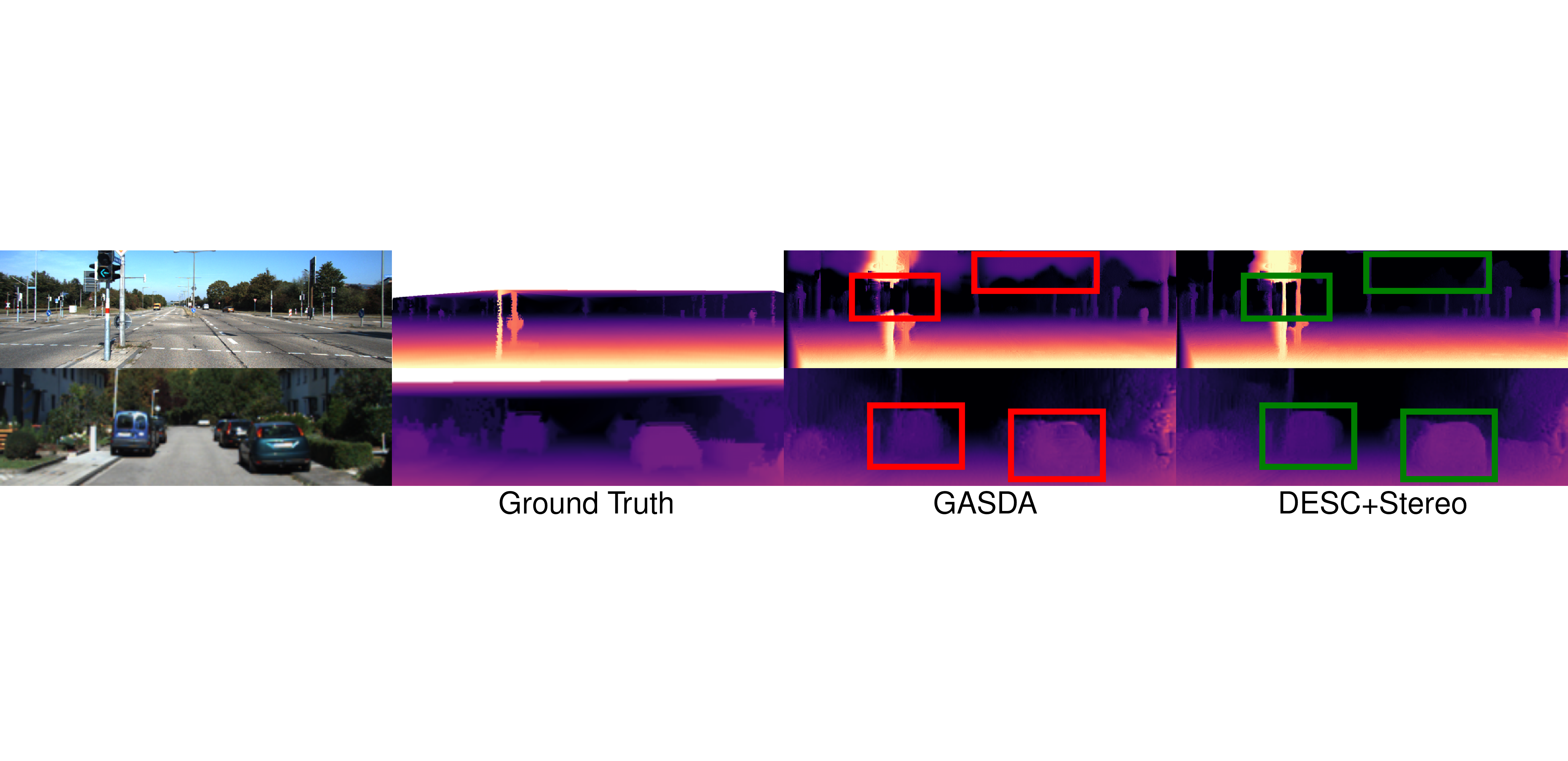}
    \vspace{-1.8em}
    \caption{Qualitative results in KITTI for models trained in Virtual KITTI$\rightarrow$KITTI with stereo supervision in KITTI. Bottom row corresponds to a center crop of the original image.}
    \label{fig:qual_res_st}
\end{figure}
\subsection{Qualitative Results }
Figure \ref{fig:qual_res} shows predictions using DESC without stereo supervision. Compared to T$^2$Net~\cite{zheng2018t2net}, we find that our method contains less high-error regions due to the guidance provided by $G_S$, as shown in the upper-right wall of the predictions in the first row. The geometry of the instances in our method tends to be complete, \eg, the cars of the second row and the larger car in the first row, which has large missing parts in the T$^2$Net prediction. Figure \ref{fig:qual_res_st} shows predictions for domain adaptation methods using stereo supervision in KITTI. Compared to GASDA, we observe a better recovery of fine structures, shown in the pole of the first row of Figure \ref{fig:qual_res_st}, and better predictions of further object instances, shown in the bottom row. DESC also predicts a better depth for the sky, as shown in the first row of Figure \ref{fig:qual_res_st}.\\
\noindent{\bf Limitations.} Due to the consistency term with $G_S$, our method shows some loss of detail in fine structures compared to T$^2$Net~\cite{zheng2018t2net}, as shown in the last row of Figure \ref{fig:qual_res}. Additionally, DESC is more computationally demanding than T$^2$Net due to the added $G_S$. Furthermore, our method relies on the quality of the computed semantic data, hence in settings where the extracted annotations are of low quality, the performance of the method may degrade.

\section{Conclusion}
We proposed a method that leverages semantic annotations to improve the performance of a depth estimation model in a domain adaptation setting. We used the relationship between instance size and depth to provide pseudo-labels in the target domain. A segmentation map and an edge map were input to a second network, whose prediction was forced to be consistent with the prediction of the main network. These additions led to higher accuracy in the settings studied. In the Virtual KITTI to KITTI benchmark, we showed a 9.8\% lower absolute relative error and a 23.6\% lower squared relative error compared to the state-of-the-art. As we use automatically extracted semantic annotations, our method can be easily added to current approaches to improve their accuracy in a domain adaptation setting, as we showed in the improvement achieved with stereo self-supervision. Approaches aiming to reduce the detail loss due to the enforced consistency of predictions could improve the method.

\noindent{\bf{Acknowledgement.}} This work was supported by UK EPSRC funding EP/S032398/1.
\bibliography{egbib}
\end{document}